\documentclass[10pt,twocolumn,letterpaper]{article}

\usepackage[pagenumbers]{cvpr} 

\usepackage{graphicx}
\usepackage{amsmath}
\usepackage{amssymb}
\usepackage{booktabs}
\usepackage{algorithm}
\usepackage{algorithmic}
\usepackage{multirow}
\usepackage{float}
\usepackage{stfloats}
\usepackage{bm}

\usepackage{newfloat}
\usepackage{listings}

\usepackage[pagebackref,breaklinks,colorlinks]{hyperref}

\usepackage[capitalize]{cleveref}
\crefname{section}{Sec.}{Secs.}
\Crefname{section}{Section}{Sections}
\Crefname{table}{Table}{Tables}
\crefname{table}{Tab.}{Tabs.}

\def\cvprPaperID{6128} 
\def\confName{CVPR}
\def\confYear{2022}

\usepackage[accsupp]{axessibility}
\begin{document}

\title{Video Shadow Detection via Spatio-Temporal Interpolation Consistency Training}

\author{Xiao Lu$^1$, Yihong Cao$^1$, Sheng Liu$^1$, Chengjiang Long$^2$, \\ 
Zipei Chen$^3$, Xuanyu Zhou$^1$, Yimin Yang$^{4,5}$, Chunxia Xiao$^{3}\thanks{Corresponding author.}$\\
$^1$College of Engineering and Design, Hunan Normal University, Changsha, China \\
$^2$Meta Reality Labs, Burlingame, CA, USA\\
$^3$School of Computer Science, Wuhan University, Wuhan, Hubei, China\\ $^4$Department of Computer Science, Lakehead University, Canada \\
$^5$Vector Institute for Artificial Intelligence, Canada\\
\tt\small $\{$luxiao,caoyihong,liusheng,zhouxy$\}$@hunnu.edu.cn, clong1@fb.com,\\
\tt\small $\{$czpp19,cxxiao$\}$@whu.edu.cn, yyang48@lakeheadu.ca}

\maketitle

\renewcommand{\thefootnote}{\fnsymbol{footnote}}

\begin{abstract}
It is challenging to annotate large-scale datasets for supervised video shadow detection methods. Using a model trained on labeled images to the video frames directly may lead to high generalization error and temporal inconsistent results. In this paper, we address these challenges by proposing a Spatio-Temporal Interpolation Consistency Training (STICT) framework to rationally feed the unlabeled video frames together with the labeled images into an image shadow detection network training. Specifically, we propose the Spatial and Temporal ICT, in which we define two new interpolation schemes, \textit{i.e.}, the spatial interpolation and the temporal interpolation. We then derive the spatial and temporal interpolation consistency constraints accordingly for enhancing generalization in the pixel-wise classification task and for encouraging temporal consistent predictions, respectively. In addition, we design a Scale-Aware Network for multi-scale shadow knowledge learning in images, and propose a scale-consistency constraint to minimize the discrepancy among the predictions at different scales. Our proposed approach is extensively validated on the ViSha dataset and a self-annotated dataset. Experimental results show that, even without video labels, our approach is better than most state of the art supervised, semi-supervised or unsupervised image/video shadow detection methods and other methods in related tasks. Code and dataset are available at \url{https://github.com/yihong-97/STICT}.
\end{abstract}

\section{Introduction}
\label{sec:intro}

\noindent Shadow detection is an important problem for many computer vision and graphics tasks, and has drawn interest in a wide range of vision applications~\cite{chen2021canet,zhang2020cla,liu2020arshadowgan,zhang2020ris}, such as object recognition~\cite{Hua:TPAMI2018,Long:IJCV2016, Long:CVPR2017, Long:ACCV2014}, virtual reality scene generation, light position estimation and object shape perception. Recently, shadow detection has achieved significant progress \cite{Zheng_19,Wang_18,Hu_17,Zhu_18,Le_18,Ding_19,chen_20} on image benchmark datasets \cite{Wang_18,Vicente_16,Zhu_10} due to the development of deep Convolutional Neural Networks (CNNs), while lacking of large-scale annotated datasets is the main reason impending the applications of deep learning-based methods in video shadow detection (VSD).

\begin{figure}[t]
\centering
\includegraphics[width=0.9\columnwidth]{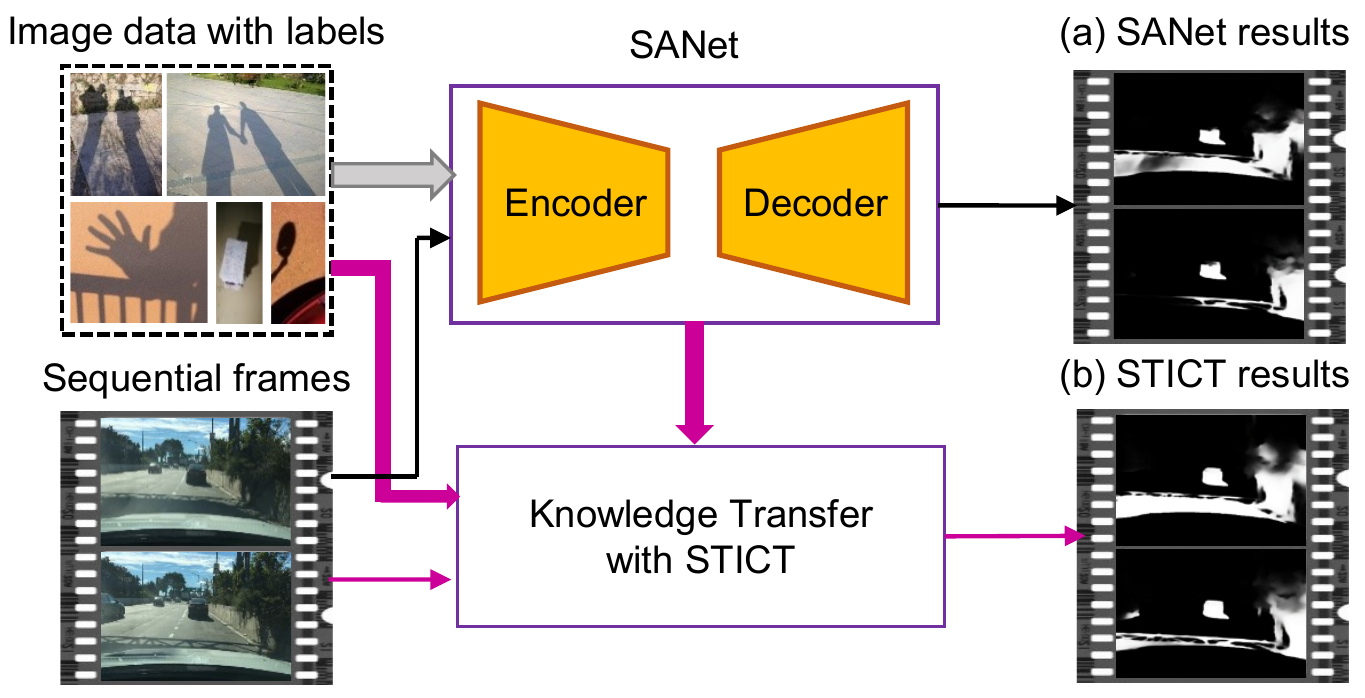} 
\caption{Shadow maps produced by our image shadow detection network SANet (a) trained on labeled images and (b) trained on both labeled images and unlabeled videos with STICT.}
\label{fig_1}
\end{figure}

How to rationally feed the unlabeled video samples into the network training, and transfer knowledge from labeled images to videos efficiently is critical for promoting the capability of deep learning-based methods on unsupervised data. However, it is rare and challenging for existing semi-supervised methods to transfer the shadow patterns in images (supervised) to videos (unsupervised) with end-to-end training, small generalization error, and temporal-consistent predictions meanwhile.

In this paper, we propose a Spatio-Temporal Interpolation Consistency Training (STICT) framework for the image-to-video shadow knowledge transfer task, in which the unlabeled video frames together with the labeled images can be rationally fed into a Scale-Aware shadow detection Network (SANet) for an end-to-end training. Accordingly, we propose a spatial interpolation consistency constraint, a temporal interpolation consistency constraint, and a scale consistency constraint to guide the network training for improving generalization, producing temporal smooth and scale consistent results. As seen from Fig.\ref{fig_1}, the detection results can be largely improved with the STICT.

To enhance the model generalization ability in our pixel-wise classification task, we propose the Spatial ICT inspired by the semi-supervised image classification method Interpolation Consistency Training (ICT) \cite{Vikas_19}. ICT encourages the prediction at a random interpolation of two unlabeled images to be consistent with the interpolation of the predictions at those two images. As proved in \cite{Vikas_19}, the samples lying near the class boundary are beneficial to enforce the decision boundary to traverse the low-density distribution regions for better generalization. Unlike the random interpolation between images in RGB space that used in \cite{Vikas_19}, we propose the spatial interpolation that is the interpolation of two uncorrelated pixels in the feature space. Our spatial interpolation is motivated by the intuitions 1) the interpolations of uncorrelated samples are more likely to locate near the class boundary to smooth the decision boundaries; 2) the interpolations of semantic pixels are more meaningful for pixel-wise classification task. Then, we derive a spatial interpolation consistency constraint accordingly to guide the network training for generalization improvement.

To encourage the temporal consistent predictions, we propose the Temporal ICT to track the prediction of the same pixel among sequential frames, in which we propose to use the temporal interpolation between two consecutive frames along the time-axis via optical flow. Then, we derive a temporal interpolation consistency constraint to guide the network training for producing temporal smooth results. Comparing with other methods that use multi-frame features or correlations among frames for temporal consistency, our method guides the network training by this extra constraint, and processes each frame independently for inference without introducing computation overhead. We would highlight that the spatial and temporal interpolations are conducted during training process, which makes our framework quite simple for inference.

Considering that the shadows in videos usually exhibit large changes in scale, we design a Scale-Aware Network (SANet) as the single-frame network for image shadow knowledge learning in the STICT framework. Unlike the traditional encoder-decoder network for shadow feature learning, SANet is designed as a encoder-decoder-refiner structure with a feature fusion module and a detail attentive module, to learn image shadow knowledge at different scales. We also propose a scale-consistency constraint accordingly to minimize the discrepancy among the predictions at different scales.

We summarize our contributions as following:

(1) We propose a STICT framework for image-to-video shadow detection task, which is rarely considered in the existing semi-supervised methods. All the labeled images and unlabeled video frames can be rationally fed into an image shadow detection network for an end-to-end training, which guarantees a compact and real-time model for inference.

(2) We propose the Spatial and Temporal ICT, in which we define two new interpolation schemes, the spatial interpolation and the temporal interpolation, for better generalization in the pixel-wise classification task and for temporal consistency, respectively. We design the SANet as the single-frame network in STICT for multi-scale shadow feature learning, and propose a scale consistency constraint accordingly for obtaining accurate shadow maps.

(3) We annotate a challenging dataset for VSD task. Experimental results on ViSha and our self-annotated dataset show that our approach performs better than most of the existing SOTA supervised/semi-supervised/unsupervised image and video methods.

\section{Related Work}

\textbf{Image Shadow Detection.} Fully-supervised deep learning-based image shadow detection \cite{Hu_17,Zheng_19,Le_18,chen_20} has recently achieved significant progress either by learning discriminative features \cite{Zheng_19,Islam:CVPR2020} or contextual information \cite{Hu_17}. Le \textit{et al.} \cite{Le_18} proposed to use GAN to generate examples with hard-to-predict cases to enhance the generalization ability. Chen \textit{et al.} \cite{chen_20} proposed a multi-task semi-supervised network to leverage unlabeled data for better generalization.

\textbf{Video Shadow Detection (VSD).} Traditional VSD methods \cite{Sanin_12,Jung_09,Liu_07,Martel_07,Huang_09} tried to identify shadow regions via statistical model\cite{Jung_09, Martel_07} by using hand-crafted features, which are sensitive to illumination change. Recently, Chen \textit{et al.} \cite{Zhu_21} annotated the first large-scale dataset (ViSha) and proposed the triple-cooperative network (TVSD-Net) for fully-supervised VSD, which utilizes triple parallel networks to learn discriminative representations at intra- and inter-video levels.

\textbf{Semi-Supervised Learning (SSL).} The assumption that the decision boundary should traverse a low-density path in the input distribution has inspired many consistency-regularization SSL methods. They vary in how to choose the data perturbation/augmentation method to encourage the invariant predictions of the unlabeled sample and its perturbed one. Some methods augment the unlabeled samples in the RGB space, \textit{e.g.}, ICT \cite{Vikas_19}, Cutmix \cite{Yun_19} and GridShuffle \cite{Chen_19}. Ouali \textit{et al.} \cite{Ouali_20} pointed out that the augmentations in RGB space is difficult to satisfy the cluster assumption for the pixel-wise classification task, and proposed to apply perturbations to the encoder's output. Our Spatial ICT is different from \cite{Ouali_20} in the perturbation method, as we use the spatial interpolation to generate new samples lying near the decision boundary for smoothing the decision boundary more efficiently than the manual perturbations.

\textbf{Temporal Consistency.} To address the temporal consistency problem, a few methods take the correlations in the video sequence into account, \textit{e.g.}, by propagating the features or results to the neighbouring frames using optical flow \cite{Nilsson_18} or recurrent unit \cite{Yan_19}, or by obtaining frame features using multi-frame information \cite{Zhu_21}, which may lead to inaccurate results or unbalanced latency. Liu \textit{et al.} \cite{Liu_20} proposed to consider the temporal consistency among frames as extra constraints during training, and process each frame independently for compact models and real-time execution. In this paper, our temporal consistency constraint is derived from the Temporal ICT, which is different from that in \cite{Liu_20}.

\begin{figure*}
\vspace{-0.75cm}
\begin{center}
\includegraphics[width=1.97\columnwidth]{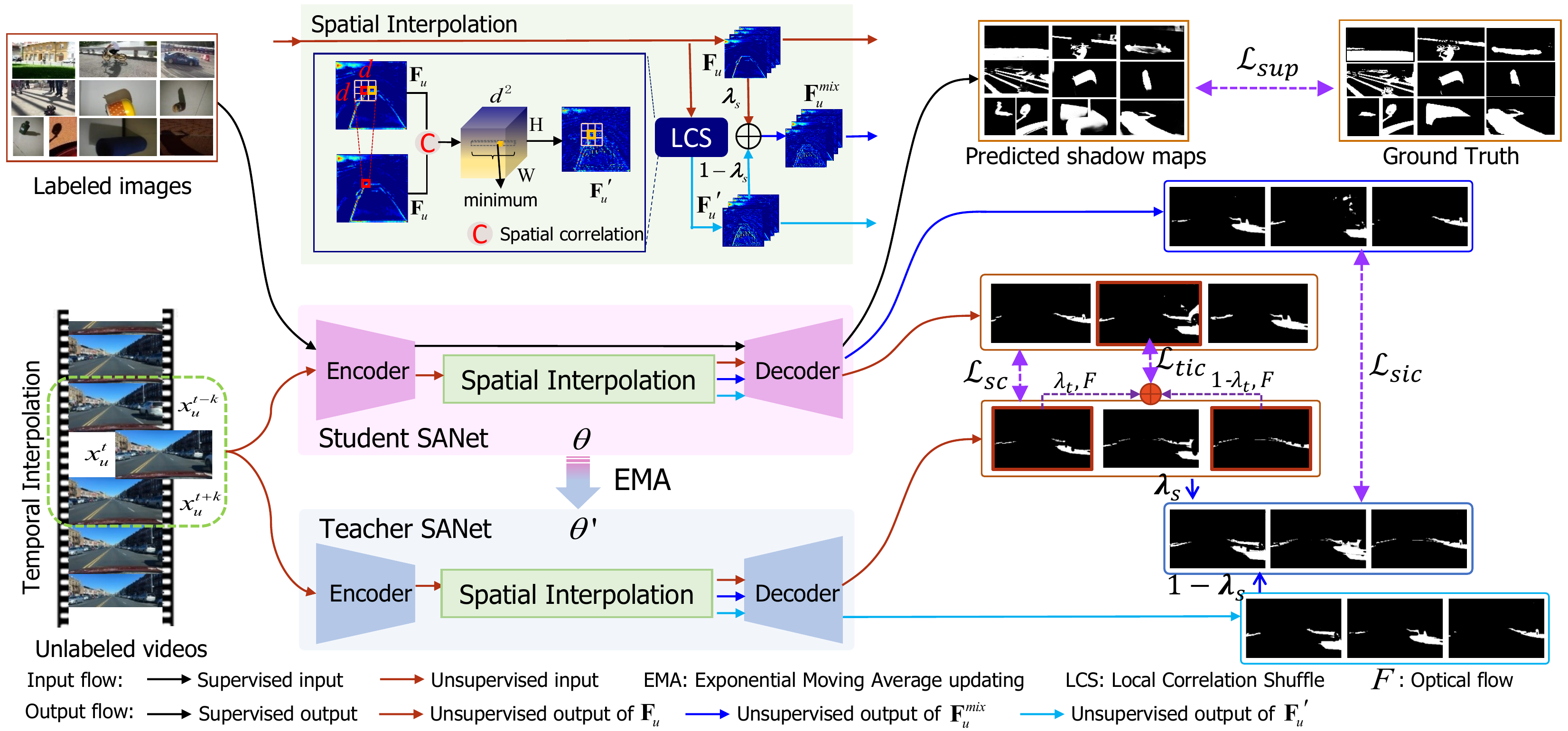}
\end{center}
\vspace{-2ex}
\caption{The overview of our method. The SANet (see Fig. \ref{fig:SANet}) is used as the student and teacher network to learn accurate shadow features. During the training phase, the labeled images are fed into the student to obtain the supervised loss $\mathcal{L}_{sup}$. Before the unlabeled frames fed into the student and teacher, temporal interpolation is conducted between two sequential frames $x_u^{t-k}$ and $x_u^{t+k}$ to generate the middle frame $x_u^{t}$. Then, each of them go through the spatial interpolation module obtaining the original feature map $\mathbf{F}_u$, the locally shuffled feature map $\mathbf{F}_u'$, and the spatial interpolation feature map $\mathbf{F}_u^{mix}$ in the student and teacher, respectively. The three consistency loss $\mathcal{L}_{sic}$, $\mathcal{L}_{tic}$ and $\mathcal{L}_{sc}$ and the $\mathcal{L}_{sup}$ are used to update the student network, while the teacher network is updated via EMA.}
\label{fig:framework}
\end{figure*}

\section{Proposed Method}

In this work, we have access to the labeled image set $\mathcal{X}_L = \{(x_l^i, y_l^i)\}_{i=1}^N$ and the unlabeled video frames $\mathcal{X}_U = \{x_u^1,\cdots, x_u^T\}$. Our method learns a VSD model by feeding the images and video frames into an image network via a Mean-Teacher learning scheme \cite{Tarvainen_17} (as seen in Fig.\ref{fig:framework}). We use the SANet as the student (parameterized by $\theta$). The teacher shares the same structure as that of the student but is parameterized by $\theta'$, $\theta'$ is an Exponential Moving Average (EMA) of $\theta$, \textit{i.e.}, $\theta_t' = \eta \theta_{t-1}' + (1 - \eta) \theta_t$. During training, the labeled images are fed into the student SANet to compute the supervised loss $\mathcal{L}_{sup}$ (Eq.\ref{eq_supLoss}). The unlabeled video frames are fed into the teacher and student simultaneously for computing the spatial interpolation consistency loss $\mathcal{L}_{sic}$ (Eq.\ref{eq_SICL}), the temporal interpolation consistency loss $\mathcal{L}_{tic}$ (Eq.\ref{eq_interLoss}) and the scale consistency loss $\mathcal{L}_{sc}$ (Eq.\ref{eq_scloss}).

The total loss for updating $\theta$ is the sum of the supervised loss and the three consistency loss:
\begin{equation}
\label{eq_Loss}
\mathcal{L}_{total} = \sum_{i=1}^{N}{\mathcal{L}_{sup}}(x_l^i) + \beta  \sum_{t=1}^T{\mathcal{L}_{cons}(x_u^t)},
\end{equation}
where $\mathcal{L}_{cons}(x_u^t)=\eta_1\mathcal{L}_{sic}(x_u^t)+\eta_2\mathcal{L}_{tic}(x_u^t)+\eta_3\mathcal{L}_{sc}(x_u^t)$, $\beta$, $\eta_1$, $\eta_2$ and $\eta_3$ are the weight parameters. The spatial and temporal interpolations are only conducted in the training phase. During the testing procedure, we only utilize the student network to predict the shadow map for each input frame independently, thus no computation overhead is introduced for inference.

\subsection{Spatial ICT}
\label{sec:SICT}

According to the cluster assumption \cite{Vikas_19}: the samples belonging to the same cluster in the input distribution are likely to belong to the same class, then it is easy to infer that comparing with the random interpolation used in \cite{Vikas_19}, the interpolations between two uncorrelated samples (possibly belonging to different classes) would be more likely to locate near the class boundary and be more useful for pushing the decision boundary far away from the class boundaries. In addition, interpolations of two images in the RGB space lacking of semantic information is meaningless for our pixel-wise classification task. Therefore, we propose the spatial interpolation, which is an interpolation of two uncorrelated pixels in the feature space. However, it is computational costly to find the most uncorrelated one for each pixel in the whole feature map. To solve this issue, we propose an easy plug-in module, called Local Correlation Shuffle (LCS) as illustrated in Fig.\ref{fig:framework}, to find the most uncorrelated pixel in the $d \times d$ local spatial area.

Let $\mathbf{F}_u \in \mathcal{R}^{H \times W \times C}$ be the feature map of $x_u$, given a location $p$ in $\mathbf{F}_u$ and $p'$ a neighborhood of it, the LCS module computes the semantic correlation by
\begin{equation}
\label{eq_corr}
c(p, p') = \mathbf{F}_u (p) \mathbf{F}_u (p')^T.
\end{equation}
The above operation will transverse the $d \times d$ area centered on $p$, and outputs a $d^2$-dimensional correlation vector. By replacing the centering pixel by the one with the minimum correlation with it, we can obtain the locally shuffled feature map $\mathbf{F}_u'$, in which each pixel is the locally most uncorrelated one of the corresponding pixel in $\mathbf{F}_u$.

Then, the spatial interpolation for all the pixels in $\mathbf{F}_u$ can be calculated as the interpolation of $\mathbf{F}_u$ and $\mathbf{F}_u'$: 
\begin{equation}
\label{eq_SICT}
\mathbf{F}_{u}^{mix} = \bm{\lambda}_s \odot \mathbf{F}_u + (1-\bm{\lambda}_s) \odot \mathbf{F}_u',
\end{equation}
where $\bm{\lambda}_s \in \mathcal{R}^{H \times W}$ with each element obeys the uniform distribution in $[0,1]$, $\odot$ is the Hadamard product on each channel of the feature map. According to the interpolation consistency constraint in ICT \cite{Vikas_19}, we derive the spatial interpolation consistent constraint as
\begin{equation}
\label{eq_SIC}
f_{\theta}(\mathbf{F}_{u}^{mix}) \approx \bm{\lambda}_s \odot f_{\theta'}(\mathbf{F}_u) + (1-\bm{\lambda}_s) \odot f_{\theta'}(\mathbf{F}_u').
\end{equation}
Accordingly, we get the spatial interpolation consistency loss that penalizes the difference between the student's and the teacher's predictions by
\begin{equation}
\label{eq_SICL}
\mathcal{L}_{sic}(x_u) = \Phi_{mse} (f_{\theta}(\mathbf{F}_u^{mix}),
\bm{\lambda}_s \odot f_{\theta'}(\mathbf{F}_u)
 + (1-\bm{\lambda}_s) \odot f_{\theta'}(\mathbf{F}_u')),
\end{equation}
where $\Phi_{mse}$ is the mean squared error loss.

\subsection{Temporal ICT}
\label{sec:TICT}

Considering the temporal changes between consecutive frames, we interpolate two consecutive frames along the time-axis and regularize the student learning to obtain temporal consistent predictions. Specifically, we interpolate the pixels in unlabeled frames $x_u^{t-k}$ and $x_u^{t+k}$ with optical flow to generate the middle frame $\hat x_u^t$,
\begin{equation}
\begin{split}
\label{eq_interOF}
\hat x_u^t &= Mix_{\lambda_t}(x_u^{t-k}, x_u^{t+k})  \\
\approx \lambda_t  g \big( x_u^{t-k}, & F_{t \to t-k} \big)+ (1 - \lambda_t)  g \big(x_u^{t+k}, F_{t \to t+k} \big),
\end{split}
\end{equation}

where $F_{t \to t-k}$ and $F_{t \to t+k}$ are optical flow from the $\hat x_u^t$ to $x_u^{t-k}$ and $x_u^{t+k}$ respectively, $g(\cdot,\cdot)$ is the differentiable bilinear interpolation function for warping a frame along the optical flow, $\lambda_t$ is a parameter controlling the contribution of two frames. Then, each pixel in $\hat x_u^t$ can be seen as the interpolation of the pixels in $x_u^{t-k}$ and $x_u^{t+k}$ along the time axis. Note that the $t_{th}$ frame $x_u^t$ is already existed in our problem, and it can be seen as the interpolation between $x_u^{t-k}$ and $x_u^{t+k}$ naturally. Then, according to Eq.\ref{eq_interOF}, we regularize the student learning by enforcing the following temporal interpolation consistent constraint
\begin{equation}
\begin{split}
\label{eq_interC}
f_{\theta}(x_u^t)=f_{\theta} & \big(Mix_{\lambda_t}(x_u^{t-k}, x_u^{t+k})\big)  \\
\approx \lambda_t g \big( f_{\theta '}(x_u^{t-k}), F_{t \to t-k} & \big)+ (1 - \lambda_t) g \big( f_{\theta '}(x_u^{t+k}), F_{t \to t+k} \big).
\end{split}
\end{equation}

Accordingly, we get the temporal interpolation consistency loss that penalizes the difference between the student's prediction $f_{\theta}(x_u^t)$ and the teacher's predictions $f_{\theta '}(x_u^{t-k})$ and $f_{\theta '}(x_u^{t+k})$, which is computed as
\begin{equation}
\label{eq_interLoss}
\begin{split}
\mathcal{L}_{tic}(x_u^t) &= \Phi_{mse} \Big( f_{\theta}(x_u^t),  \lambda_t g \big( f_{\theta '}(x_u^{t-k}), F_{t \to t-k}\big) \\
& + (1 - \lambda_t) g\big( f_{\theta '}(x_u^{t+k}), F_{t \to t+k}\big) \Big),
\end{split}
\end{equation}
where $\lambda_t$ is set to be 0.5 in this paper, and $F_{t \to t-k}$ and $F_{t \to t+k}$ can be calculated via a pre-trained optical flow prediction network (\textit{i.e.}, FlowNet2.0 \cite{Eddy_17}).

\subsection{SANet and Scale Consistency Constraint}
\label{sec:frameCNN}

\begin{figure}[ht]
\begin{center}
\includegraphics[width=0.49\textwidth]{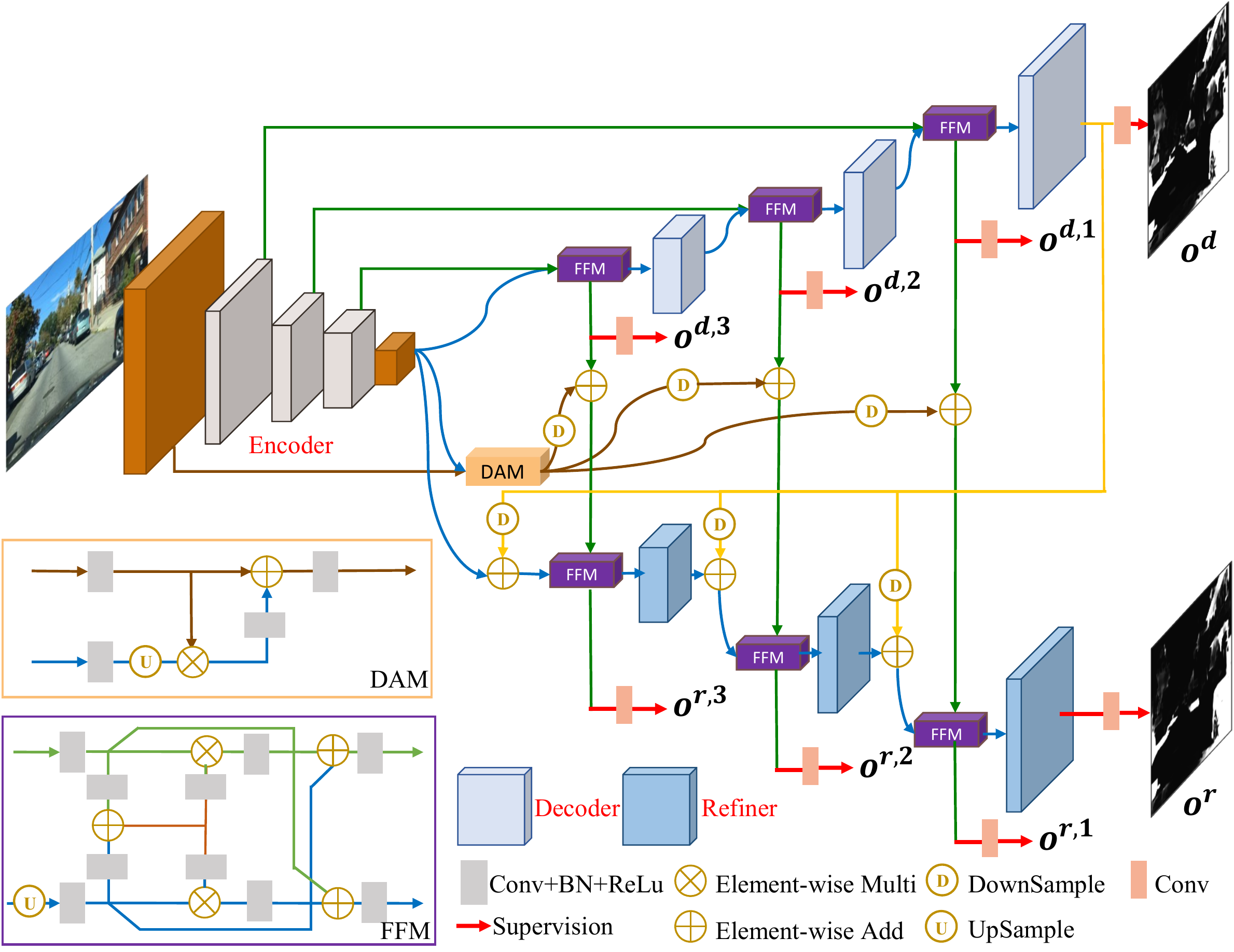}
\end{center}
\vspace{-1ex}
   \caption{The architecture of our SANet.}
\label{fig:SANet}
\end{figure}

\textbf{Scale-Aware Network (SANet).} Traditional methods use the Encoder-Decoder structure to aggregate multi-level features to produce the final shadow map. However, features of different level may have missing and distorted parts due to downsamplings, upsamplings and noises. To maintain the semantic features, complement with the missing details and refine the distorted features, we design the SANet with the Feature Fusion Module (FFM), the Detail Attentive Module (DAM) and the Encoder-Decoder-Refiner (EDR) structure to obtain accurate shadow maps at multi-scale.

The FFM is designed to keep the semantic information and enrich the details meanwhile. As seen in Fig.\ref{fig:SANet}, FFM firstly fuses the high-level semantic features (the blue input branch) and low-level detail features (the green input branch) by element-wise adding, then it takes the selective fusion strategy by element-wise multiplication to focus on the semantic and detail features, at high and low level respectively. Lastly, the high- and low-level features are complemented with each other by an element-wise adding. Compared with traditional fusion strategy, FFM can highlight the semantic features with details.

Addressing the feature distortion and missing caused by down/upsampling, we propose to add a refiner and feed the features before the last convolution layer in the decoder back to the refiner (the yellow branch) for features correcting and refining. Therefore, comparing with the top-down process in the decoder, the refiner has both the top-down and bottom-up process. In the top-down process, multi-level features are aggregated via FFM gradually, and in the bottom-up process, the aggregated features are fed back to each level for refining.

DAM firstly fuses the highest and lowest level features selectively element-wise multiplication to highlight the details with semantic information, then the fused features are complemented with the lowest level features, which are further fed into the refiner for enhancing details.

We use the above three modules to get an accurate prediction finally ($o^r$ in Fig.\ref{fig:SANet}). With deep supervision at multi-level, we obtain eight outputs at three scales for an image. As shown in Fig.\ref{fig:SANet}, $o^{d,1} \sim o^{d,3}$ and $o^{r,1} \sim o^{r,3}$ are the multi-scale outputs of decoder and refiner, respectively. $o^{d}$ and $o^{r}$ are the final outputs of the decoder and refiner, respectively.
We use the pixel position aware loss $\mathcal{L}_{ppa}$ \cite{Wei_20}, which is the sum of the weighted binary cross entropy loss and the weighted IoU loss, for supervising each output. The whole supervised loss is defined as
\begin{equation}
\label{eq_supLoss}
\mathcal{L}_{sup} =  \sum_{i=1}^3 \frac{1}{2^i}\big(\mathcal{L}_{ppa}^{d,i} + \mathcal{L}_{ppa}^{r,i} \big)  + \frac{1}{2} \big( \mathcal{L}_{ppa}^{d}+\mathcal{L}_{ppa}^{r} \big),
\end{equation}
where $\mathcal{L}_{ppa}^{d,i}$ and $\mathcal{L}_{ppa}^{r,i}$ are the loss on the $i_{th}$ scale of decoder and refiner, respectively, $\frac{1}{2^{i}}$ is the weight of output at different scales, and $\mathcal{L}_{ppa}^{d}$ and $\mathcal{L}_{ppa}^{r}$ are the loss on the final output of decoder and refiner, respectively.

\textbf{Scale consistency constraint.} To reduce the influence of noise on the output of each scale, we introduce the scale consistency constraint by minimizing the discrepancy (\textit{i.e.}, variance) among the predictions at different scales. For each unlabeled image $x_u$, we use the refiner's outputs $o^{r,1} \sim o^{r,3}$ as the multi-scale results, then we get the teacher's and student's predictions $\{f_{\theta'}^s(x_u)\}_{s=1}^3$ and $\{f_{\theta}^s(x_u)\}_{s=1}^3$. To reduce the variance among the predictions at different scales, we propose to minimize the difference between the teacher's average prediction and that of the student at all scales. The teacher's average prediction is denoted by
\begin{equation}
\label{eq_avg}
f_{\theta'}^{ave}(x_u)= \frac{1}{3}\sum_{s=1}^3 f_{\theta'}^{s}(x_u).
\end{equation}
Then, the scale consistency loss is defined as
\begin{equation}
\label{eq_scloss}
\mathcal{L}_{sc}(x_u) = \frac{1}{3}\sum_{s=1}^3 \Phi_{mse}(f_{\theta}^s(x_u), f_{\theta'}^{ave}(x_u)).
\end{equation}

\subsection{Implementation Details}

We initialize the backbone of SANet by ResNet-50 \cite{He_16}, to accelerate the training procedure and reduce the over-fitting risk, and other parameters in SANet are initialized as random values. The Adam algorithm is used to optimize the student network with maximum learning rate $0.0003$ for ResNet-50 backbone and $0.003$ for other parts. We also adopt a linear decay strategy to update learning rate. The mini-batch size is set to be 4. Moreover, instead of mixing labeled and unlabeled samples together in a mini-batch as that in \cite{chen_20}, we process mini-batches from the source and target datasets separately so that batch normalization uses different normalization statistics for each domain during training. The decay parameter $\eta$ in EMA is empirically set as 0.999. For the consistency loss weight $\beta$, according to \cite{Tarvainen_17}, we use the Gaussian ramp-up function for updating: $\beta(t) = \beta_{max}e^{-5(1-t/t_{max})^2}$, and $t_{max}=10$, $\beta_{max}$ is set to be $1$. The consistency loss parameters are set to be $\eta_1 =\eta_2 =\eta_3= 1$, and the parameter $k$ in Eq.\ref{eq_interLoss} is set to be 1. Our implementation is developed using PyTorch and all the experiments are conducted on a single NVIDIA GTX 2080Ti GPU, and it takes about 15ms for our method to predict the shadow map for a frame.

\section{Experiments}
\label{sec:exp}
\renewcommand{\thefootnote}{\arabic{footnote}}
\textbf{Datasets.} We use the training set in the image shadow detection benchmark dataset SBU \cite{Wang_18} as the labeled images, and transfer the shadow patterns to ViSha \cite{Zhu_21} and our self-collected VIdeo ShAdow Detection dataset (VISAD) for evaluating our method\footnote{The authors Xiao Lu, Yihong Cao, and Sheng Liu signed the non-commercial licenses, downloaded the datasets, and produced all the experimental
results in this paper. Meta didn't have access to all these datasets.}. The SBU dataset is the largest annotated image shadow dataset with natural scenes, including 4,089 training images and 638 testing images. ViSha is the first dataset for VSD, which contains 120 videos with 11,685 frames, and we use the same data partitioning as that in  \cite{Zhu_21}. VISAD consists of 81 videos belonging to the BDD-100K \cite{Yu_20}, DAVSOD \cite{fan2019_DAVSOD}, DAVIS \cite{Perazzi_16} and FBMS \cite{Brox_10,ochs2013seg}, and we divide them into two parts according to 
the scenes: the Driving Scenes (VISAD-DS) part and the Moving Object Scenes (VISAD-MOS) part, denoted as DS and MOS, respectively. We annotate the cast shadow manually in 33 videos densely with the LabelImg\footnote{https://pypi.org/project/labelImg/}. Some details about DS and MOS are presented in Table \ref{tab:data}.

\vspace{-1ex}
\begin{table}[ht]\scriptsize
\begin{center}
\begin{tabular}{c|c|c|c|c}
\hline
                         & SCD  & $\sharp$ V / $\sharp$AV & $\sharp$ F / $\sharp$AF  & Resolution\\
\hline\hline
DS                       & BDD   & 47 / 17     & 7,953 / 2,881 & 1280 $\times$ 720 \\
\hline
\multirow{3}{*}{MOS}     & DAVIS & 15 / 15     & 1,047 / 1,047 & (540$\sim$1920) $\times$ (394$\sim$1080) \\
                         & DAVSOD& 9 /  0      & 1,134 / 0 & 640 $\times$ 360 \\
                         & FBMS  & 10 / 1      & 2,432 / 260 & (530$\sim$960) $\times$ (360$\sim$540) \\
\hline
\end{tabular}
\end{center}
\vspace{-2ex}
\caption{Some details about our VISAD Dataset. SCD: source dataset. $\sharp$V: number of videos. $\sharp$AV: number of annotated videos. $\sharp$F: number of frames. $\sharp$AF: number of annotated frames. }
\label{tab:data}
\end{table}

\noindent \textbf{Evaluation Metric.} Following recent works on shadow detection, we employ the balance error rate (\textbf{BER}) which considers both of the detection quality of shadow and non-shadow regions, to quantitatively evaluate the shadow detection performance. In addition, we follow \cite{Zhu_21} to employ three other metric that commonly used in salient object detection, the Mean Absolute Error (\textbf{MAE}), F-measure ($\mathbf{F}_{\beta}$) and Intersection over Union (\textbf{IoU}), to evaluate the performance. In general, a smaller BER and MAE, and a larger ${F}_{\beta}$ and IoU indicate a better detection performance.

\subsection{Ablation study}

We start the ablation study on the important modules to better understand their behavior and effectiveness on shadow knowledge learning and transferring. Our ablation study is conducted on DS as the scenes in DS are very different from that in the SBU, and it is more difficult to transfer knowledge to DS than to the other two datasets.

\vspace{-2ex}
\begin{figure}[H]
\begin{center}
\includegraphics[width=0.48\textwidth]{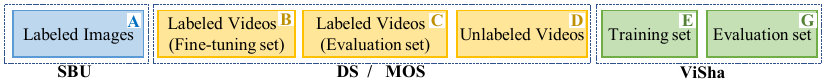}
\end{center}
\vspace{-3ex}
\caption{Dataset partition in our experiments.}
\label{fig:data}
\end{figure}
\vspace{-2ex}

\begin{figure*}[ht]
\begin{center}
\includegraphics[width=0.98\textwidth]{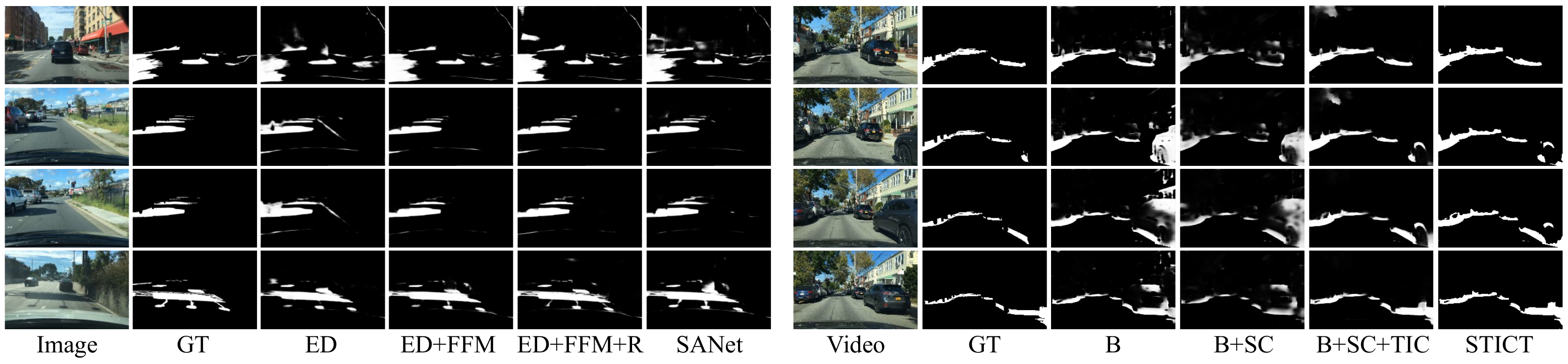}
\end{center}
\vspace{-2ex}
   \caption{Visualization results of our ablation study. The left part: ablation results on SANet. The right part: ablation results on STICT.}
\label{fig:vis_ablation}
\end{figure*}

\textbf{Ablation study on SANet.} To evaluate the three proposed modules, EDR, FFM and DAM in the SANet, we consider three baseline networks. The first baseline is constructed by using only the encoder-decoder structure (\textbf{ED}), with a simple fusing structure (upsample+add+conv) replacing the FFM structure. The second is to add the FFM structure in the decoder (\textbf{ED+FFM}), while the third is to add the refiner structure in the network (\textbf{ED+FFM+R}). Finally, we consider our SANet with the DAM structure. As seen the data partition in Fig.\ref{fig:data}, all the networks are trained on A and fine-tuned on B, and are tested on C. The upper part of Table \ref{tab:ablation_result1} summarizes the results of our SANet and three baseline networks. From the results, we can see that FFM has the most effectiveness on promoting the IoU value, while the refiner structure is most effective on $F_{\beta}$. Although DAM is very simple, it is very important and necessary for learning accurate shadow knowledge in images. The visualization results presented in the left part of Fig.\ref{fig:vis_ablation} demonstrate the effectiveness of each module on the detection of details and small scale shadow regions.

\begin{table}[ht]\scriptsize
\begin{center}
\begin{tabular}{c|c|c|c|c|c|c|c}
\hline
\cline{1-8}
   ED                & FFM          & R          & DAM        & MAE$\downarrow$ & $F_{\beta}$ $\uparrow$  & IoU $\uparrow$     & BER$\downarrow$    \\
\hline
$\checkmark$         &              &             &             & 0.035          & 0.569          & 0.433           & 15.29\\
$\checkmark$         & $\checkmark$ &             &             & 0.031          & 0.616          & 0.492           & 13.48\\
$\checkmark$         & $\checkmark$ &$\checkmark$ &             & 0.029          & 0.660          & \textbf{0.514}  & 13.21\\
$\checkmark$         & $\checkmark$ &$\checkmark$ &$\checkmark$ & \textbf{0.028} & \textbf{0.708} & \textbf{0.514}  & \textbf{13.14}\\
\hline\hline
                     &SC           &  TIC       &  SIC         & MAE $\downarrow$ & $F_{\beta}$ $\uparrow$ & IoU $\uparrow$ & BER $\downarrow$\\
\hline
\multirow{4}{*}{B}   &              &             &               & 0.093            & 0.501                  & 0.304          & 17.01\\
                     &$\checkmark$  &             &               & 0.092            & 0.518                  & 0.311          & 16.78\\
                     &$\checkmark$  &$\checkmark$ &               & 0.079            & 0.587                  & 0.320          & 16.29\\
                     &$\checkmark$  &$\checkmark$ &$\checkmark$   & \textbf{0.065}   & \textbf{0.646}         & \textbf{0.370} &\textbf{14.17}\\
\hline\hline
\multicolumn{2}{c|}{Space} & \multicolumn{2}{c|}{Scheme} & MAE $\downarrow$ & $F_{\beta}$ $\uparrow$ & IoU $\uparrow$ & BER $\downarrow$\\
\hline
\multicolumn{2}{c|}{RGB space} & \multicolumn{2}{c|}{RI} & 0.072 & 0.447 & 0.350 & 14.80 \\
\cline{1-8}
\multicolumn{2}{c|}{\multirow{2}{*}{Feature space}} & \multicolumn{2}{c|}{RI} & 0.068 & 0.557 & 0.356 & 15.24 \\
\cline{3-8}
\multicolumn{2}{c|}{ }&  \multicolumn{2}{c|}{SI} &\textbf{0.065}&\textbf{0.646}&\textbf{0.370}&\textbf{14.17}\\
\hline
\end{tabular}
\end{center}
\vspace{-2ex}
\caption{The upper part: ablation results on SANet pretrained on SBU and fine-tuned on DS, R: Refiner. The middle part: ablation results on STICT, B: basic SANet trained on SBU without fine-tuning on DS. The lower part: ablation results on interpolation schemes, RI: random interpolation. SI: spatial interpolation.}
\label{tab:ablation_result1}
\end{table}

\textbf{Ablation study on three consistency constraints.} We consider four baseline methods, the first method is to apply the SANet trained on SBU to the target videos directly (denoted as \textbf{B}). Then we train three models by adding the three consistency constraints, scale consistency (\textbf{SC}), temporal interpolation consistency (\textbf{TIC}) and spatial interpolation consistency (\textbf{SIC}), to the basic model sequentially. As seen the data partition in Fig.\ref{fig:data}, all the networks are trained on A and D, and are tested on C. The results are reported in the middle part of Table \ref{tab:ablation_result1}. As we can see, the temporal consistency constraint has significant boosting performance on reducing MAE and improving $F_{\beta}$ values, while the spatial consistency constraint has positive effect on improving IoU and reducing BER values. Visualization results presented in the right part of Fig.\ref{fig:vis_ablation} also confirm with that of the quantitative results, which demonstrates that all components are necessary for the proposed framework for accurate and temporal consistent shadow maps.

\textbf{Comparison with other interpolation schemes in Spatial ICT.} We also compare the performance of our spatial interpolation (\textbf{SI}) in the feature space with that of the random interpolation (\textbf{RI}) in the feature space and RGB space, respectively. The results in the lower part of Table \ref{tab:ablation_result1} show that interpolation in the feature space is more meaningful than in the RGB space, and our spatial interpolation is effective for the pixel-wise classification task, which also demonstrates the effectiveness of our LCS module.

\begin{table*}[ht]\footnotesize
\begin{center}
\begin{tabular}{c|c|c|c|c|c|c|c|c|c|c|c|c|c}
\hline
\multirow{2}{*}{}       & \multirow{2}{*}{Method}  &  \multicolumn{4}{c|}{ViSha}     &  \multicolumn{4}{c|}{DS}      &  \multicolumn{4}{c}{MOS}  \\
\cline{3-14}
                        &   & MAE $\downarrow$ & $F_{\beta}$ $\uparrow$ & IoU $\uparrow$ & BER $\downarrow$ & MAE $\downarrow$ & $F_{\beta}$ $\uparrow$ & IoU $\uparrow$ & BER $\downarrow$ & MAE $\downarrow$ & $F_{\beta}$ $\uparrow$ & IoU $\uparrow$ & BER $\downarrow$  \\
\hline\hline
\multirow{8}{*}{I.S.}   & DSC\cite{Hu_17}         & 0.096 & 0.514         & 0.434         & 17.91  & 0.096& 0.507& 0.315& 18.24 & 0.070& 0.573& 0.385& 24.18 \\
                        & BDRAR\cite{Zhu_18}       & 0.050 & 0.695         & 0.484         & 21.29  & 0.088& 0.504& 0.284& 15.25 & 0.130& 0.456& 0.250& 18.79 \\
                        & DSDNet\cite{Zheng_19}          & 0.044 & 0.702         & 0.518         & 19.88  & 0.068& 0.408& 0.301& 18.42 & 0.083& 0.595& 0.365& 19.62 \\
                        & MTMT-SSL\cite{chen_20}     & 0.080 & 0.664         & 0.500         & 18.11  & 0.106& 0.521& 0.298& 19.49 & 0.085& 0.575& 0.402& 25.61 \\
                        & ECANet\cite{fang2021robust}     & \textbf{0.032} & 0.741         & 0.539         & 20.06  & 0.037& 0.583& 0.379& 23.67 & 0.078& 0.565& 0.336& 28.68 \\
                        & FSDNet\cite{hu2021revisiting}     & 0.047 & 0.681         & 0.473         & 22.86  & 0.029& 0.623& 0.377& 27.77 & 0.084& \textbf{0.634}& 0.359& 28.38 \\
                        & MagNet\cite{huynh2021progressive}     & 0.045 & 0.685        & 0.507        & 20.41  & 0.038& 0.606& 0.399& 21.56 & 0.080& 0.586& 0.341& 28.91 \\
                        & SANet(Ours)& 0.036 & 0.752         & \textbf{0.596}& \textbf{13.26}  & \textbf{0.028}& \textbf{0.708}& \textbf{0.514}& 13.14 & 0.091& 0.601& 0.341& 25.93 \\
\hline
I.U.                    & MTMT-Uns.\cite{chen_20}    & 0.088 & 0.568         & 0.457         & 20.10  & 0.154& 0.309  & 0.232& 20.19 & 0.081& 0.564& 0.391& 27.01 \\
\hline
\multirow{3}{*}{V.S.}   & TVSD-Net\cite{Zhu_21}     & 0.033& \textbf{0.757} & 0.567& 17.70  & 0.032  & 0.634  & 0.508  & \textbf{11.55} & 0.191& 0.313& 0.227& 20.24 \\
                        & GRFP\cite{Nilsson_18}         & 0.059 & 0.682 & 0.531  & 20.64  & 0.057& 0.611  & 0.326 & 18.87  & 0.115  & 0.551 & 0.292& 26.76 \\
                        & NS\cite{Liang_20}           & 0.061 & 0.586 & 0.405  & 24.17  & 0.073& 0.495  & 0.339 & 19.16  & 0.115  & 0.534 & 0.261& 29.60 \\
\hline
\multirow{2}{*}{V.U.}   & RCRNet\cite{Yan_19}       & 0.093 & 0.490 & 0.346  & 26.89  & 0.067& 0.377  & 0.236 & 28.76  & 0.088  & 0.596 & 0.356 & 20.13\\
                        & STICT(Ours)  & 0.046 & 0.702& 0.545  & 16.60  & 0.065  & 0.646& 0.370  & 14.17& \textbf{0.058}& 0.625& \textbf{0.409}& \textbf{18.51} \\
\hline
\end{tabular}
\end{center}
\caption{Comparison results of our method with SOTA methods. I.S.: Image-based supervised methods. I.U.: Image-based method without labels. V.S.: Video-based supervised methods. V.U.: Video-based method without labels. The best results are highlighted with bold.}
\label{tab:sets_result}
\end{table*}

\subsection{Comparison with SOTA Methods}

We make comparison with several SOTA image/video based methods on shadow detection and other relative tasks, including six supervised image shadow detection methods DSC \cite{Hu_17}, BDRAR \cite{Zhu_18}, DSDNet \cite{Zheng_19}, ECANet \cite{fang2021robust} and FSDNet\cite{hu2021revisiting}, a semi-supervised method MTMT \cite{chen_20}, a supervised VSD method TVSD-Net \cite{Zhu_21}, two video semantic segmentation methods GRFP \cite{Nilsson_18}, NS \cite{Liang_20}, an image semantic segmentation method MagNet \cite{huynh2021progressive}, and a video salient object detection method RCRNet \cite{Yan_19} to demonstrate the effectiveness of our method.

To make the comparisons fair, the data partition is shown in Fig.\ref{fig:data}. For the image- and video-based supervised methods (\textbf{I.S.}/\textbf{V.S.}), we train the models on A, and fine-tune them on B (or labeled E) by reducing the learning rates and iterations to $1/10$ of those in their published papers. For the image- and video-based methods without video labels (\textbf{I.U.}/\textbf{V.U.}), we train the models on A and D (or unlabeled E). For training both of GRFP and NS, we use PSPNet \cite{Zhao_17a} as the backbone, and pre-train it on SBU to get the single-frame prediction. Then, we follow the training procedure as given in the published paper \cite{Nilsson_18} and \cite{Liang_20}, and use the training set and the fine-tuning set for training the whole network while keep other settings unchanged. For training RCRNet \cite{Yan_19} in the unsupervised scenario, we use SBU for training RCRNet, and the produced shadow maps are used as the sparse labels to generate the pseudo labels, and then train the RCRNet+NER model for results refinement.

\textbf{Quantitative comparisons.} Table \ref{tab:sets_result} summarizes the quantitative results of different methods on the three groups of experiments. From the results on ViSha and DS, we can see that the performance of STICT ranks third to our supervised SANet and the supervised method TVSD-Net, but it has better performance on BER value and $F_{\beta}$ value than TVSD-Net on ViSha and DS, respectively. Moreover, STICT performs much better than other methods in MOS, and it has $29.3\%$ and $10.3\%$ lower MAE and BER score, and $6.5\%$ higher IoU score comparing to the second best-performing method. The main reason for the performance gap between the DS and MOS is that the scenes are very similar to each other in DS, and a small number of labels can make the supervised model fit to the whole test set. However, the scenes in MOS are different from each other, then a large amount of unlabeled data can make our model generalize to the test set.

Although our method is not the best one in all the three datasets, it has more stable performance compared with other methods, because it has a good way to adapt to different domains for improving generalization ability. Since we train our model without annotations on the video datasets, the competitive and even superior performance on these three datasets validates the generalization capability of our proposed algorithm. Besides, it should be noted that our supervised SANet performs the best on DS and it beats all other competitors on IoU and BER values on ViSha, which demonstrates that our SANet is effective for detection the shadows in multi-scales.

\begin{figure*}[ht]
\vspace{-0.5cm}
\begin{center}
\includegraphics[width=0.98\textwidth]{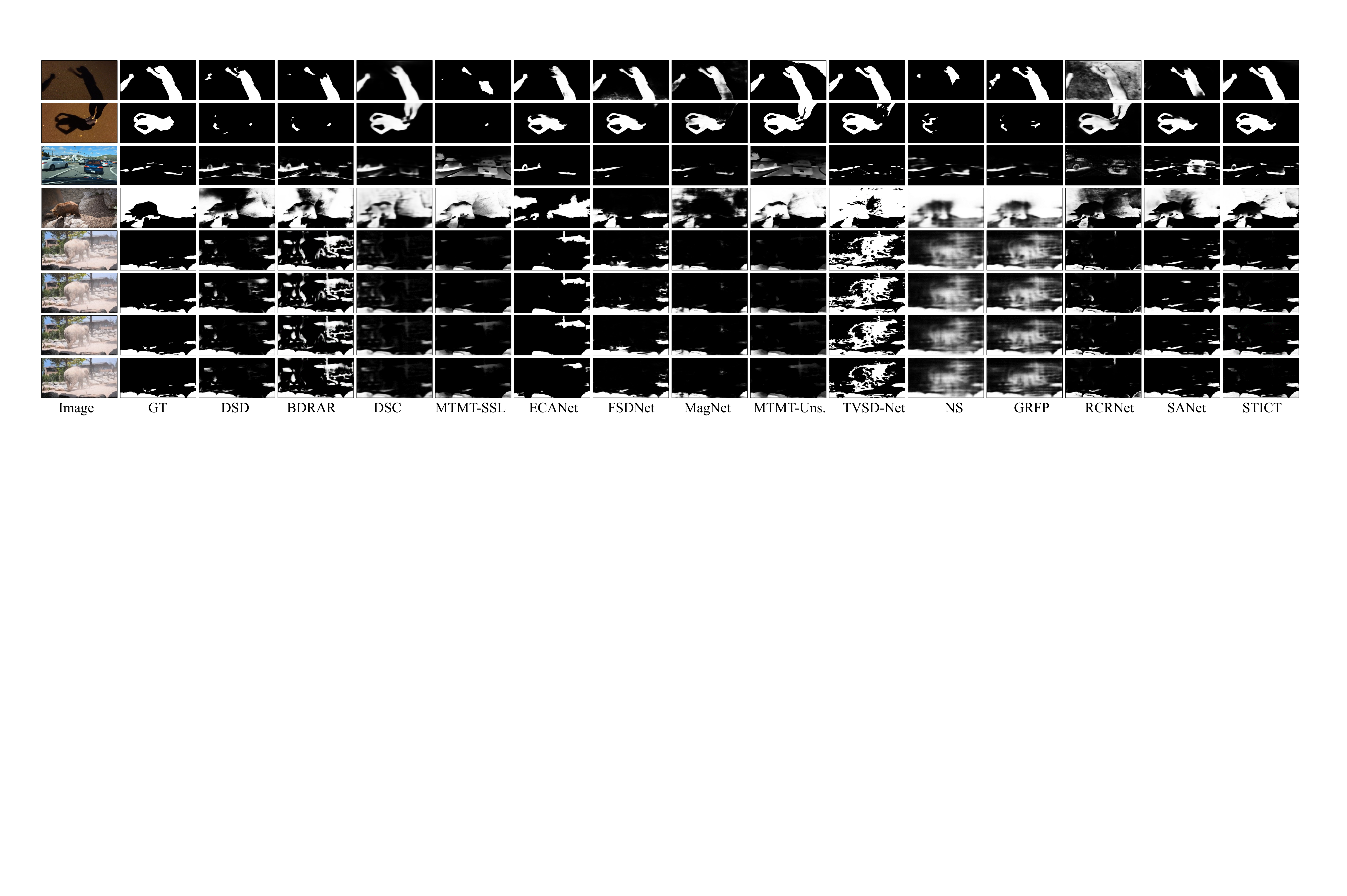}
\end{center}
\vspace{-2ex}
   \caption{Visual comparison of shadow maps produced by our method and other comparative methods.}
\label{fig:viscom_img}
\vspace{-0.2cm}
\end{figure*}

\textbf{Qualitative comparisons.} We further visually compare the shadow maps produced by our method and the competitors in Fig. \ref{fig:viscom_img}. From the images and corresponding labels, we can see some challenging cases are included: the shadows in low illumination condition (the 1st row), the small scale shadows (the 3rd row), shadows around complex backgrounds (the 4th row), bright object in shadow regions (the 2nd row) and the dark objects around shadows (the 3rd row). It is obvious that our STICT and our supervised SANet can locate the various scale shadow regions and discriminate the shadow details from its complex background accurately. While other methods tend to misrecognize the dark non-shadow regions as shadows and neglect the small scale regions. The results in the lower part of Fig. \ref{fig:viscom_img} also verify that our method can produce temporal consistent predictions. All the visual comparisons demonstrate the efficiency and generalization capability of our method.

\begin{figure}[ht]
\begin{center}
\includegraphics[width=0.45\textwidth]{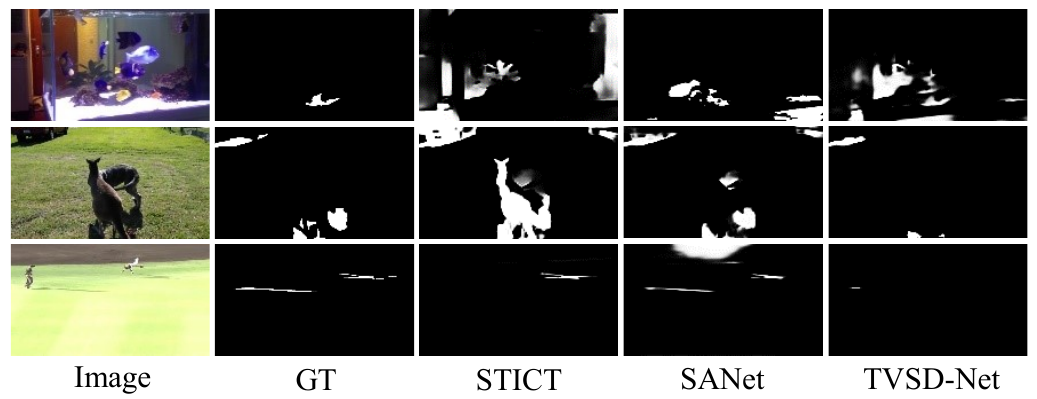}
\end{center}
\vspace{-2ex}
   \caption{Some failure cases of our method.}
\label{fig:failure_case}
\end{figure}

\textbf{Limitations.} From the failure cases of our method presented in Fig. \ref{fig:failure_case}, we can see that our method has failed in the scenes with large illumination contrast (the 1st row), it sometimes misrecognize the self-shadow regions (the 2nd row), and it can not detect the soft shadow effectively (the 3rd row). However, our supervised SANet performs much better in the above cases. This performance gap results from the difference between the source image dataset and the target video dataset: 1) it lacks of the shadow knowledge in large illumination contrast scene and soft-shadow pattern in SBU, 2) some of the self-shadow regions are annotated in SBU, while they are not annotated in the video dataset.

\section{Conclusions}
\label{sec:con}

We have presented a novel VSD method by transferring shadow knowledge from the labeled images to the unlabeled videos. The key idea is to feed the labeled images and unlabeled video frames into the SANet training via the STICT framework, which enhances the generalization ability and encourages the temporal and scale consistent predictions. Experimental results of our method are better than those of most supervised, semi-supervised or unsupervised image/video methods.

\textbf{Societal impact.} VSD has a positive effect on computer vision, but improper use of it may result in potential negative impacts, \textit{e.g.}, violation modification of evidence videos.

\section{Acknowledgement}
\label{sec:ack}

This work is partially supported by Prof. Xiao Lu's NSFC (No. 62007007), Prof. Chunxia Xiao's NSFC (No. 61972298) and Prof. Chunxia Xiao's Key Technological Innovation Projects of Hubei Province (2018AAA062).


{\small
\bibliographystyle{ieee_fullname}
\bibliography{camera_ready_v.bbl}

\begin{thebibliography}{10}\itemsep=-1pt

\bibitem{Brox_10}
Thomas Brox and Jitendra Malik.
\newblock Object segmentation by long term analysis of point trajectories.
\newblock In {\em European conference on computer vision}, pages 282--295.
  Springer, 2010.

\bibitem{Liang_20}
Liang~Chieh Chen, Raphael~Gontijo Lopes, Bowen Cheng, Maxwell~D. Collins,
  Ekin~D. Cubuk, Barret Zoph, Hartwig Adam, and Jonathon Shlens.
\newblock Naive-student: Leveraging semi-supervised learning in video sequences
  for urban scene segmentation.
\newblock {\em European Conference on Computer Vision (ECCV)}, 2020.

\bibitem{Chen_19}
Yue Chen, Yalong Bai, Wei Zhang, and Tao Mei.
\newblock Destruction and construction learning for fine-grained image
  recognition.
\newblock In {\em Proceedings of the IEEE/CVF Conference on Computer Vision and
  Pattern Recognition}, pages 5157--5166, 2019.

\bibitem{chen2021canet}
Zipei Chen, Chengjiang Long, Ling Zhang, and Chunxia Xiao.
\newblock Canet: A context-aware network for shadow removal.
\newblock In {\em Proceedings of the IEEE/CVF International Conference on
  Computer Vision}, pages 4743--4752, 2021.

\bibitem{Zhu_21}
Zhihao Chen, Liang Wan, Lei Zhu, Jia Shen, Huazhu Fu, Wennan Liu, and Jin Qin.
\newblock Triple-cooperative video shadow detection.
\newblock In {\em 2021 IEEE Conference on Computer Vision and Pattern
  Recognition (CVPR)}, 2021.

\bibitem{chen_20}
Zhihao Chen, Lei Zhu, Liang Wan, Song Wang, and Pheng~Ann Heng.
\newblock A multi-task mean teacher for semi-supervised shadow detection.
\newblock In {\em IEEE/CVF Conference on Computer Vision and Pattern
  Recognition (CVPR 2020)}, 2020.

\bibitem{Ding_19}
Bin Ding, Chengjiang Long, Ling Zhang, and Chunxia Xiao.
\newblock Argan: Attentive recurrent generative adversarial network for shadow
  detection and removal.
\newblock {\em Proceedings of the IEEE/CVF International Conference on Computer
  Vision}, pages 10213--10222, 2019.

\bibitem{fan2019_DAVSOD}
Deng-Ping Fan, Wenguan Wang, Ming-Ming Cheng, and Jianbing Shen.
\newblock Shifting more attention to video salient object detection.
\newblock In {\em Proceedings of the IEEE/CVF Conference on Computer Vision and
  Pattern Recognition}, pages 8554--8564, 2019.

\bibitem{fang2021robust}
Xianyong Fang, Xiaohao He, Linbo Wang, and Jianbing Shen.
\newblock Robust shadow detection by exploring effective shadow contexts.
\newblock In {\em Proceedings of the 29th ACM International Conference on
  Multimedia}, pages 2927--2935, 2021.

\bibitem{He_16}
Kaiming He, Xiangyu Zhang, Shaoqing Ren, and Jian Sun.
\newblock Deep residual learning for image recognition.
\newblock In {\em IEEE/CVF Conference on Computer Vision and Pattern
  Recognition}, 2016.

\bibitem{hu2021revisiting}
Xiaowei Hu, Tianyu Wang, Chi-Wing Fu, Yitong Jiang, Qiong Wang, and Pheng-Ann
  Heng.
\newblock Revisiting shadow detection: A new benchmark dataset for complex
  world.
\newblock {\em IEEE Transactions on Image Processing}, 30:1925--1934, 2021.

\bibitem{Hu_17}
Xiaowei Hu, Lei Zhu, Chi~Wing Fu, Jing Qin, and Pheng~Ann Heng.
\newblock Direction-aware spatial context features for shadow detection.
\newblock {\em IEEE Conference on Computer Vision and Pattern Recognition
  (CVPR)}, 2018.

\bibitem{Hua:TPAMI2018}
Gang Hua, Chengjiang Long, Ming Yang, and Yan Gao.
\newblock Collaborative active visual recognition from crowds: A distributed
  ensemble approach.
\newblock {\em IEEE Transactions on Pattern Analysis and Machine Intelligence
  (T-PAMI)}, 40(3):582--594, 2018.

\bibitem{Huang_09}
Jia-Bin Huang and Chu-Song Chen.
\newblock Moving cast shadow detection using physics-based features.
\newblock In {\em 2009 IEEE Conference on Computer Vision and Pattern
  Recognition}, pages 2310--2317. IEEE, 2009.

\bibitem{huynh2021progressive}
Chuong Huynh, Anh~Tuan Tran, Khoa Luu, and Minh Hoai.
\newblock Progressive semantic segmentation.
\newblock In {\em Proceedings of the IEEE/CVF Conference on Computer Vision and
  Pattern Recognition}, pages 16755--16764, 2021.

\bibitem{Eddy_17}
Eddy Ilg, Nikolaus Mayer, Tonmoy Saikia, Margret Keuper, and Thomas Brox.
\newblock Flownet 2.0: Evolution of optical flow estimation with deep networks.
\newblock In {\em IEEE/CVF Conference on Computer Vision and Pattern
  Recognition}. IEEE, 2017.

\bibitem{Islam:CVPR2020}
Ashraful Islam, Chengjiang Long, Arslan Basharat, and Anthony Hoogs.
\newblock Doa-gan: Dual-order attentive generative adversarial network for
  image copy-move forgery detection and localization.
\newblock In {\em Proceedings of the IEEE Conference on Computer Vision and
  Pattern Recognition}, 2020.

\bibitem{Jung_09}
Cláudio~Rosito Jung.
\newblock Efficient background subtraction and shadow removal for monochromatic
  video sequences.
\newblock {\em IEEE Transactions on Multimedia}, 11(3):571--577, 2009.

\bibitem{Le_18}
Hieu Le, Tomas F.~Yago Vicente, Vu Nguyen, Minh Hoai, and Dimitris Samaras.
\newblock A+d net: Training a shadow detector with adversarial shadow
  attenuation.
\newblock {\em Proceedings of the European Conference on Computer Vision
  (ECCV)}, 2018.

\bibitem{liu2020arshadowgan}
Daquan Liu, Chengjiang Long, Hongpan Zhang, Hanning Yu, Xinzhi Dong, and
  Chunxia Xiao.
\newblock Arshadowgan: Shadow generative adversarial network for augmented
  reality in single light scenes.
\newblock In {\em Proceedings of the IEEE/CVF Conference on Computer Vision and
  Pattern Recognition}, pages 8139--8148, 2020.

\bibitem{Liu_20}
Yifan Liu, Chunhua Shen, Changqian Yu, and Jingdong Wang.
\newblock Efficient semantic video segmentation with per-frame inference.
\newblock In {\em European Conference on Computer Vision}, pages 352--368.
  Springer, 2020.

\bibitem{Liu_07}
Zhou Liu, Kaiqi Huang, Tieniu Tan, and Liangsheng Wang.
\newblock Cast shadow removal combining local and global features.
\newblock In {\em 2007 IEEE Conference on Computer Vision and Pattern
  Recognition}, pages 1--8. IEEE, 2007.

\bibitem{Long:CVPR2017}
Chengjiang Long and Gang Hua.
\newblock Correlational gaussian processes for cross-domain visual recognition.
\newblock In {\em Proceedings of the IEEE Conference on Computer Vision and
  Pattern Recognition (CVPR)}, pages 118--126, 2017.

\bibitem{Long:IJCV2016}
Chengjiang Long, Gang Hua, and Ashish Kapoor.
\newblock A joint gaussian process model for active visual recognition with
  expertise estimation in crowdsourcing.
\newblock {\em International Journal of Computer Vision (IJCV)},
  116(2):136--160, 2016.

\bibitem{Long:ACCV2014}
Chengjiang Long, Xiaoyu Wang, Gang Hua, Ming Yang, and Yuanqing Lin.
\newblock Accurate object detection with location relaxation and regionlets
  re-localization.
\newblock In {\em The 12th Asian Conference on Computer Vision (ACCV)}, pages
  3000--3016. IEEE, 2014.

\bibitem{Martel_07}
Nicolas Martel-Brisson and Andre Zaccarin.
\newblock Learning and removing cast shadows through a multidistribution
  approach.
\newblock {\em IEEE Transactions on Pattern Analysis and Machine Intelligence},
  29(7):1133--1146, 2007.

\bibitem{Nilsson_18}
David Nilsson and Cristian Sminchisescu.
\newblock Semantic video segmentation by gated recurrent flow propagation.
\newblock In {\em IEEE/CVF Conference on Computer Vision and Pattern
  Recognition}. IEEE, 2018.

\bibitem{ochs2013seg}
Peter Ochs, Jitendra Malik, and Thomas Brox.
\newblock Segmentation of moving objects by long term video analysis.
\newblock {\em IEEE transactions on pattern analysis and machine intelligence},
  36(6):1187--1200, 2013.

\bibitem{Ouali_20}
Yassine Ouali, Céline Hudelot, and Myriam Tami.
\newblock Semi-supervised semantic segmentation with cross-consistency
  training.
\newblock In {\em Proceedings of the IEEE/CVF Conference on Computer Vision and
  Pattern Recognition}, pages 12674--12684, 2020.

\bibitem{Perazzi_16}
Federico Perazzi, Jordi Pont-Tuset, Brian McWilliams, Luc Van~Gool, Markus
  Gross, and Alexander Sorkine-Hornung.
\newblock A benchmark dataset and evaluation methodology for video object
  segmentation.
\newblock In {\em Proceedings of the IEEE conference on computer vision and
  pattern recognition}, pages 724--732, 2016.

\bibitem{Sanin_12}
Andres Sanin, Conrad Sanderson, and Brian~C Lovell.
\newblock Shadow detection: A survey and comparative evaluation of recent
  methods.
\newblock {\em Pattern recognition}, 45(4):1684--1695, 2012.

\bibitem{Tarvainen_17}
Antti Tarvainen and Harri Valpola.
\newblock Mean teachers are better role models: Weight-averaged consistency
  targets improve semi-supervised deep learning results.
\newblock {\em Conference and Workshop on Neural Information Processing Systems
  (NIPS)}, 2017.

\bibitem{Vicente_16}
Tomos F~Yago Vicente, Le Hou, Chen-Ping Yu, Minh Hoai, and Dimitris Samaras.
\newblock Large-scale training of shadow detectors with noisily-annotated
  shadow examples.
\newblock In {\em European Conference on Computer Vision}, pages 816--832.
  Springer, 2016.

\bibitem{Vikas_19}
Verma Vikas, Kawaguchi Kenji, Lamb Alex, Kannala Juho, Bengio Yoshua, and
  Lopez-Paz David.
\newblock Interpolation consistency training for semi-supervised learning.
\newblock In {\em Proceedings of the Twenty-Eighth International Joint
  Conferenceon Artificial Intelligence (IJCAI-19)}, 2019.

\bibitem{Wang_18}
Jifeng Wang, Xiang Li, and Jian Yang.
\newblock Stacked conditional generative adversarial networks for jointly
  learning shadow detection and shadow removal.
\newblock In {\em Proceedings of the IEEE Conference on Computer Vision and
  Pattern Recognition}, pages 1788--1797, 2018.

\bibitem{Wei_20}
Jun Wei, Shuhui Wang, and Qingming Huang.
\newblock F3net: Fusion, feedback and focus for salient object detection.
\newblock In {\em Proceedings of the AAAI Conference on Artificial
  Intelligence}, 2020.

\bibitem{Yan_19}
Pengxiang Yan, Guanbin Li, Yuan Xie, Zhen Li, Chuan Wang, Tianshui Chen, and
  Liang Lin.
\newblock Semi-supervised video salient object detection using pseudo-labels.
\newblock In {\em Proceedings of the IEEE/CVF International Conference on
  Computer Vision}, pages 7284--7293, 2019.

\bibitem{Yu_20}
Fisher Yu, Haofeng Chen, Xin Wang, Wenqi Xian, Yingying Chen, Fangchen Liu,
  Vashisht Madhavan, and Trevor Darrell.
\newblock Bdd100k: A diverse driving dataset for heterogeneous multitask
  learning.
\newblock {\em Proceedings of the IEEE/CVF conference on computer vision and
  pattern recognition}, pages 2636--2645, 2020.

\bibitem{Yun_19}
Sangdoo Yun, Dongyoon Han, Seong~Joon Oh, Sanghyuk Chun, Junsuk Choe, and
  Youngjoon Yoo.
\newblock Cutmix: Regularization strategy to train strong classifiers with
  localizable features.
\newblock In {\em Proceedings of the IEEE/CVF International Conference on
  Computer Vision}, pages 6023--6032, 2019.

\bibitem{zhang2020cla}
Ling Zhang, Chengjiang Long, Qingan Yan, Xiaolong Zhang, and Chunxia Xiao.
\newblock Cla-gan: A context and lightness aware generative adversarial network
  for shadow removal.
\newblock In {\em Computer Graphics Forum}, volume~39, pages 483--494. Wiley
  Online Library, 2020.

\bibitem{zhang2020ris}
Ling Zhang, Chengjiang Long, Xiaolong Zhang, and Chunxia Xiao.
\newblock Ris-gan: Explore residual and illumination with generative
  adversarial networks for shadow removal.
\newblock In {\em Proceedings of the AAAI Conference on Artificial
  Intelligence}, volume~34, pages 12829--12836, 2020.

\bibitem{Zhao_17a}
H. Zhao, J. Shi, X. Qi, X. Wang, and J. Jia.
\newblock Pyramid scene parsing network.
\newblock In {\em 2017 IEEE Conference on Computer Vision and Pattern
  Recognition (CVPR)}, 2017.

\bibitem{Zheng_19}
Quanlong Zheng, Xiaotian Qiao, Ying Cao, and Rynson W.~H. Lau.
\newblock Distraction-aware shadow detection.
\newblock In {\em 2019 IEEE/CVF Conference on Computer Vision and Pattern
  Recognition (CVPR)}, 2019.

\bibitem{Zhu_10}
Jiejie Zhu, Kegan~GG Samuel, Syed~Z Masood, and Marshall~F Tappen.
\newblock Learning to recognize shadows in monochromatic natural images.
\newblock In {\em 2010 IEEE Computer Society conference on computer vision and
  pattern recognition}, pages 223--230. IEEE, 2010.

\bibitem{Zhu_18}
Lei Zhu, Zijun Deng, Xiaowei Hu, Chi~Wing Fu, Xuemiao Xu, Jing Qin, and
  Pheng~Ann Heng.
\newblock Bidirectional feature pyramid network with recurrent attention
  residual modules for shadow detection.
\newblock In {\em European Conference on Computer Vision (ECCV)}, 2018.

\end{thebibliography}
}

\end{document}